# Title - A Subspecialty Diagnostic System Powered by Self-Supervised Learning for Cervical Histopathology

## Author Information

Yizhi Wang[1†], Li Chen[1†], Qiang Huang[1,2†], Tian Guan[1†], Xi Deng[3], Zhiyuan Shen[4], Jiawen Li[1], Xinrui Chen[1], Bin Hu[5], Zhe Lei[6], Xitong Ling[1], Taojie Zhu[1], Zirui Huang[1], Deshui Yu[1], Fanyi Zeng[1],Yan Liu[7], Jiurun Chen[1,8], Lianghui Zhu[1], Qiming He[1,9], Yiqing Liu[10], Diwei Shi[1,11], Hanzhong Liu[12], Junbo Hu[13], Hongyi Gao[14], Zhen Song[1,8], Xilong Zhao[15,16], Lingchuan Guo[6*], Chao He[17*], Ming Zhao[3*], Yonghong He[1*]

**Affiliations:**

[1]Shenzhen International Graduate School, Tsinghua University; Shenzhen, China.

[2]Shenzhen Shengqiang Technology Co., Ltd.; Shenzhen, China.

[3]Ningbo Clinical Pathology Diagnosis Center; Ningbo, China.

[4]Greater Bay Area National Center of Technology Innovation; Guangzhou, China.

[5]Department of Pathology, Affiliated Jinhua Hospital, Zhejiang University School of Medicine; Jinhua, Zhejiang Province, China.

[6]Department of Pathology, the First Affiliated Hospital of Soochow University, Institute of Clinical Pathology and Precision Medicine of Soochow University, Suzhou 215006, Jiangsu, China.

[7]Hangzhou Adicon Clinical Laboratories Co., Ltd.; Hangzhou, Zhejiang Province, China.

[8]Department of Network Intelligence, Pengcheng Laboratory; Shenzhen, 518055, China.

[9]Interdisciplinary Institute for Medical Engineering, Fuzhou University; Fuzhou, 350108, China

[10]Jinfeng Laboratory; Chongqing. China.

[11]Research Institute of Tsinghua University in Shenzhen; Shenzhen, China.

[12]Xiaogan Central Hospital, Wuhan University of Science and Technology; Xiaogan, Hubei Province, China.

[13]Department of Pathology, Maternal and Child Health Hospital of Hubei Province; Wuhan, China.

[14]Department of Pathology, Guangdong Women and Children Hospital; Guangzhou, China.

[15]The Basic Medical Laboratory of the 920th Hospital of the Joint Logistics Support Force of PLA; Kunming, Yunnan 650032, China.

[16]The Integrated Engineering Research Center of Cell Biological Medicine of State and Regions; Kunming, Yunnan 650032, China.

[17]Department of Engineering Science, University of Oxford; Oxford, UK.

*Corresponding author. Email:

Lingchuan Guo: szglc@hotmail.com;

Chao He: chao.he@eng.ox.ac.uk;

Ming Zhao: zhaomingpathol@163.com;

**Yonghong He: heyh@sz.tsinghua.edu.cn;**

† These authors contributed equally to this work

## Abstract

Cervical cancer remains a major malignancy in women, with large-scale human papillomavirus-based screening generating a large volume of biopsies and, for advanced cases, surgical specimens, imposing an overwhelming diagnostic burden that necessitates precise histopathological evaluation. Although deep learning shows promise, general foundation models offer a broader reach but remain limited in capturing subspecialty-specific features and task adaptability. We introduce the Cervical Subspecialty Pathology (CerS-Path) diagnostic system, built upon the subspecialized dataset and developed through two synergistic pretraining stages: self-supervised learning on approximately 190 million tissue patches to build a cervical-specific feature extractor, and multimodal enhancement with 2.5 million image-text pairs, followed by diagnostic functions integration and downstream task enhancement. CerS-Path integrates eight functions—across morphological diagnosis, predictive analytics, and multimodal tasks—to cover the diagnostic workflow from biopsies to resection specimens. Comprehensive evaluations demonstrate improved performance in cervical pathology in female patients, where CerS-Path outperforms state-of-the-art foundation models across 26 of 30 diagnostic tasks. Furthermore, in multicenter retrospective and prospective validation across six independent cohorts of women, CerS-Path achieves up to 99% sensitivity for clinically relevant lesion detection with acceptable specificity across sites, highlighting its potential for screening-derived histopathology and suitability for integration into high-volume diagnostic pathways.

## Introduction

Cervical cancer ranks as the fourth most common cancer among women worldwide and is largely preventable through organized screening and identification of high-grade precancerous lesions (CIN2/3/HSIL) and timely histopathological diagnosis and treatment[1,2]. First-line screening methods such as HPV testing, colposcopy, and liquid-based cytology serve as auxiliary diagnostic tools to enhance the efficiency of cervical abnormality screening[3, 4, 5]. Women with abnormal screening results are referred for colposcopy and cervical biopsy as second-line screening. Histopathological examination of these biopsies by an experienced pathologist remains the diagnostic gold standard in this workflow[6], and all cases must be reviewed with high accuracy to avoid overtreatment of LSIL/CIN1 and undertreatment of HSIL. The large volume of cervical second-line screening, combined with the demand for diagnostic precision, makes histopathological assessment a complex and labor-intensive task, placing a considerable burden on clinical pathologists[7]. Furthermore, accurate tumor grading and subtyping are essential for treatment decisions [8, 9, 10], but inter-observer variability can introduce subjectivity and affect outcomes. These factors collectively underscore an urgent clinical need for a cervical histopathology diagnostic tool that enables both efficient second-line screening and accurate diagnosis across complex real-world scenarios.

This unmet clinical need has prompted increasing interest in artificial intelligence (AI)-based computational pathology solutions. Recent advances in machine learning and deep learning have demonstrated the potential to perform diagnostic tasks in cervical histopathology [11, 12], including subtype classification [13, 14], tumor grading[15, 16, 17], lymph node metastasis prediction[18], and biomarker inference[19]. However, many existing approaches remain limited by narrow functionality, constrained generalizability, and heavy reliance on manual annotations, which hinder their clinical adoption[13, 17, 18, 20]. Effectively addressing the full spectrum of diagnostic challenges in cervical histopathology requires a comprehensive and disease-specialized diagnostic system.

The emergence of foundation models in pathology is reshaping the analytical paradigm of computational pathology, paving the way for the development of cervical subspecialty diagnostic systems. Such analytical approaches leverage self-supervised learning to shift feature extractors from task-specific training paradigms toward domain-general architectures [20, 21], thereby enhancing the potential and upper bound of AI-assisted diagnosis in histomorphological interpretation, prognostic prediction, and biomarker analysis[22, 23]. However, while general pathology foundation models offer broad applicability, they are constrained by inherent biases in their pretraining data distribution and lack the fine-grained discriminative power required to achieve the diagnostic accuracy of gynecologic pathologists, specialists in cervical histopathology[24, 25]. Achieving such accuracy demands models that are specifically trained and fine-tuned on cervical tissue data. Furthermore, contemporary diagnostic systems must address an escalating clinical requirement to handle complex multimodal clinical tasks. Existing multimodal pathology encoders, utilized as feature extractors, demonstrate critical limitations in task-specific adaptability and cross-modal semantic reasoning, particularly for cross-modal interaction tasks[26, 27, 28, 29]. With the advent of multimodal large language models (MLLMs), there is growing potential to address such challenges in a unified manner[26, 27]. To address the diverse demands of cervical subspecialty practice, diagnostic systems need not only high diagnostic precision but also the ability to support semantic interpretation, descriptive reporting, and zero-shot inference, enabling comprehensive assistance in multimodal clinical scenarios. Faced with the diverse clinical demands of cervical subspecialty diagnosis, systems must ensure high diagnostic precision while also supporting semantic-level reasoning, descriptive capabilities in multimodal scenarios, and zero-shot diagnostic inference—enabling comprehensive and context-aware clinical assistance.

To address these limitations, in this work, we developed CerS-Path (Fig. 1), a cervical histopathology diagnostic system grounded in subspecialty-enhanced representation learning and designed to meet the dual challenge of diagnostic precision and clinical complexity. The system's construction was founded on the CerS-140K dataset (~ 140,000 cervical WSIs), comprising 115,000 slides for foundational pre-training (yielding 190 million patches) and 25,000 slides dedicated to downstream evaluation and clinical-grade performance enhancement. We then performed large-scale self-supervised pretraining on this dataset, yielding CerS-V, a specialized vision encoder for cervical histomorphology. Subsequent multimodal augmentation enhanced CerS-V using 2.5 million image-text pairs, yielding CerS-M, a cross-modal embedding model optimized for cervical pathology comprehension. The final phase involved task-specialized adaptation through fine-tuning across 30 downstream diagnostic tasks aligned with cervical pathology subspecialty requirements (Fig. 2). By integrating these task-specific modules via subspecialized diagnostic logic, CerS-Path delivers eight clinical functions: screening (i.e., second-line screening), grading, subtyping, quantitative analysis, rare cancer detection, predictive modeling, multimodal Q&A, and automated reporting. This integrated diagnostic framework establishes a solution for cervical histopathology subspecialization.

Besides its comprehensive functions, CerS-Path also exhibits superior diagnostic performance. In retrospective evaluation spanning 30 diverse cervical pathology tasks across eight diagnostic functions, CerS-Path outperforms existing state-of-the-art (SOTA) models in 26 tasks (Fig. 3a), with an average improvement of 3.66 percentage points (pp) across the primary evaluation metrics for each task and an average 31.93% improvement in log-odds ratios for critical tasks (Fig. 3b, c). Furthermore, the system demonstrates advanced capabilities beyond conventional models, including anomaly detection (+21.88 pp), rare cancer classification (+4.05 pp), and molecular prediction (+5.59 pp) functions. CerS-Path also supports multimodal diagnostic tasks, expanding the application scope of foundation models in subspecialty pathology. Multicenter validation confirms the system's clinical robustness, consistently achieving sensitivity exceeding 99.0% for lesion screening, 98.6% for precancer screening, and 98.9% for cancer subtyping, alongside an F1-Score of 80.87% in the challenging Silva classification. Comprehensive evaluations show that CerS-Path offers broad diagnostic capability and deep specialization in cervical pathology, supporting the transition of general-purpose foundation models into clinically deployable, specialty-specific diagnostic systems.

# Results

## Development of the CerS-Path Diagnostic System

Anchored in the subspecialty clinical diagnostic paradigm, we developed the CerS-Path system to deliver comprehensive support throughout the cervical pathology workflow, from population-level screening to specimen diagnosis post-radical hysterectomy. The system's construction involved four integrated stages:

(I) Large-scale cervical-specific visual self-supervised pretraining. As shown in Fig. 2a, we first established the cervical-specific dataset CerS-140K, comprising about 190 million tissue patches curated from about 115,000 histopathology WSIs. Implementing multi-path contrastive learning through the DINOV2 framework [28], we pretrained a general vision transformer backbone on this dataset to derive a cervical-optimized visual feature extractor, namely CerS-V, with the detailed training framework described in the Methods section. This large-scale domain-specific pretraining enabled CerS-V to specifically enhance the characterization of cervical subsites and disease patterns while retaining generic histopathological representation capacities.

(II) Multimodal representation expansion. Building upon the expert-level visual features of CerS-V, we implemented parameter-efficient multimodal alignment with about 2.5 million histopathology image-text pairs via the CLIP [29] framework and yield CerS-M (Fig. 2b). This phase established cross-modal projection layers that transform histopathological signatures into clinically contextualized semantic spaces, while preserving the visual encoder's diagnostic fidelity through extra LoRA modules, where the detailed training diagram is described in the Methods section.

(III) Diagnostic reasoning enhancement. To augment multimodal clinical reasoning capabilities, particularly for pathology Q&A, we implemented instruction tuning using expert-curated diagnostic dialogues (Fig. 2b). Integrated with the Qwen2.5 multimodal decoder [27], this process enabled CerS-M to dynamically translate histopathological features into clinically interpretable language. The resulting CerS-M embedding model achieved domain-optimized interpretation for cervical pathology through cross-modal alignment.

(IV) Downstream module specialization and clinical integration. To bridge feature extractor capabilities with clinical workflows, we first applied task-specific fine-tuning across ROI-level, WSI-level and hybrid diagnostic paradigms, adapting the model to the spectrum of cervical pathology tasks (Fig. 2c). Subsequently, diagnostic functions were integrated through clinical reasoning pathways structured by diagnostic intent: screening and rare cancer detection for primary triage; grading, subtyping and quantitative analysis for disease characterization; risk modeling and biomarker prediction for prognostic care; alongside multimodal Q&A and automated reporting for interpretive support, with both the comprehensive integration logic and the categorization of diagnostic tasks across functions detailed in Supplementary Fig. 1.

## Performance Overview of CerS-Path

Herein, we conducted a comprehensive evaluation of CerS-Path across a broad spectrum of cervical histopathology tasks, closely aligned with real-world diagnostic scenarios. As illustrated in Fig. 3a, CerS-Path outperformed UNI[20] in 26 out of 30 diagnostic tasks (Supplementary Table 1), achieving an average performance increment of 3.66 pp. Further quantification via log-odds ratio analysis (Fig. 3b) revealed substantial improvements exceeding 31.93% (Supplementary Table 2) in critical tasks, including rare adenocarcinoma (ADC-Rare) and silva pattern subtyping (ADC-Silva). Fig. 3c demonstrates CerS-Path's superior foundational performance compared to other SOTA models, achieving a substantial margin over even the high-performing Virchow2 framework[30]. Comparisons of more general pathology models (e.g., Phikon, CtransPath, etc.) are provided in Supplementary Table 3. These results underscore the superiority of subspecialty-pretrained models like CerS-Path in cervical pathology, outperforming general-purpose counterparts in specialized tasks and enabling capabilities beyond their reach.

We quantified the scaling effects of pretraining data volume on diagnostic performance. As shown in Fig. 3d, we observed that both routine and complex diagnostic tasks reached a performance plateau at approximately 200 million image patches, establishing this as a suitable scale for downstream deployment. The scaling law demonstrates that subspecialty-guided data augmentation enhances diagnostic capability, particularly for clinically critical yet challenging tasks like ADC-Silva pattern recognition. Given consistent performance plateaus beyond 190 million tiles, we empirically selected the model pretrained on 190 million tiles as the optimal backbone for downstream deployment. Notably, UNI-based initialization provided CerS-Path with essential histopathological representations, showing superior stability and cross-task reliability compared to general-purpose ImageNet weights (Supplementary Table 4).

We further identified critical capabilities enabling subspecialty models to address specialized diagnostic challenges. As demonstrated in Fig. 3e, CerS-Path exhibits strong few-shot adaptation performance, showcasing superior transferability and robustness in both benign tissue classification and adenocarcinoma-HPV subtyping tasks. This enhanced generalization capability, surpassing general-purpose foundation models, underscores its improved adaptability for disease-specific diagnosis within subspecialty contexts. Visualization of feature space characteristics (Fig. 3f) reveals substantially improved class separability in CerS-Path compared to general pathology models, quantitatively evidenced by a lower Davies-Bouldin Index (DBI, 1.7235 vs. 2.3301 for UNI) and a higher silhouette coefficient (0.2490 vs. 0.1810 for UNI), with more comparison metrics presented in Supplementary Fig. 2 and Supplementary Table 5.

Under the ARPL open-set detection framework[31], CerS-Path achieves greater prediction confidence separation between in-domain and out-of-domain samples relative to benchmark foundation models (Fig. 3g). This discriminative capacity further enables superior identification of unknown pathological abnormalities, establishing state-of-the-art outlier detection performance (Fig. 3h).

Another notable advantage of the subspecialty model lies in its superior performance on long-tailed, domain-specific rare cancer subtyping tasks. As shown in Fig. 3i, the model achieves clearer separation of tail classes (GEAC and other rare cancers) in the t-SNE embedding space. Quantitative analyses in Supplementary Table 6 further support this, with the Davies–Bouldin Index, cosine distance, and Wasserstein distance in the original feature space all indicating enhanced separability between tail and head classes under sub-specialty optimization.

Multimodal integration significantly enhanced adaptability to complex subspecialty diagnostic workflows. As shown in Fig. 3j, we evaluated CerS-Path's zero-shot classification capability through text-guided prompting, demonstrating 76.64% (70.84% for CerS-M) diagnostic B-ACC in distinguishing benign versus malignant cervical tissue categories across diverse clinical samples (Supplementary Fig. 3) and outperforming existing pathology foundation models under identical conditions (UNI: 60.74%; Virchow2: 66.02%). The results demonstrate CerS-Path's balanced sensitivity across diverse subtypes to reduce false negatives and generalization ability to rare or unannotated cases through vision-language alignment. This paradigm reduces reliance on large, labeled datasets and supports the transition of pathology AI toward open-ended, interactive decision support.

In summary, comprehensive benchmarking underscores the role of subspecialty-enhanced pretraining in computational pathology. CerS-Path systematically outperformed general-purpose foundation models, demonstrating that domain specialization preserves broad histopathological comprehension while actively improving fundamental diagnostic metrics, such as high-precision diagnosis accuracy and consistency in challenging grading tasks. Furthermore, this subspecialty enhancement facilitates advanced clinical capabilities less evident in the evaluated general models, including out-of-distribution anomaly detection, accurate diagnosis of long-tail rare cancers, and zero-shot generalization—illustrating a crucial transition from general feature representation to rigorous, clinical-grade diagnostic utility.

## Evaluation of CerS-Path on Core Visual Diagnostic Tasks

In the retrospective study, we evaluated the performance of the CerS-Path system across four core diagnostic functions: screening, grading, subtyping, and quantitative analysis. This evaluation utilized the downstream part of CerS-140K collected from the Ningbo Pathology Center (NBPC) and covered 18 routine visual tasks, including 4 ROI-level tasks and 14 WSI-level tasks. Quantitative comparisons demonstrated superior balanced accuracy (B-ACC) in both WSI analysis (Fig. 4a) and ROI-level tasks (Fig. 4b).

Specifically, CerS-Path achieved clinically significant performance in two pivotal second-line screening benchmarks: 93.09% (95% CI: 92.29%-93.89%) B-ACC for lesion detection (Task Screen-Lesion, lesioned vs. benign), and 98.99% (95% CI: 98.22%-99.76%) B-ACC in cancer subtype classification (Cancer Subtype). This performance level holds significant population health implications, as large-scale screening can reduce cervical cancer incidence and mortality by 60%-80%[5, 11, 32]. As evidenced by the five-fold cross-validated ROC curves in Fig. 4c, the system demonstrated superior sensitivity, with B-ACC improvements of 1.85 pp ($P = 0.0086$ ) and 0.88 pp ($P = 0.038$) over UNI, respectively. Moreover, Fig. 4d demonstrates the capability of CerS-Path in detecting subtle lesions, while Fig. 4e highlights its ability to distinguish micro-lesions of cervical cancer between Adenocarcinoma (ADC) and squamous cell carcinoma (SCC), with high-attention regions indicating the key discriminative features for this differentiation.

Accurate morphological grading of squamous intraepithelial lesions (SIL) and Silva grading of glandular lesions are clinically essential for assessing disease severity and guiding therapeutic decisions[2, 33, 34]. As evidenced by the confusion matrices in Fig. 4f and Fig. 4g, CerS-Path achieved strong consistency in classic grading tasks, reaching 78.62% (95% CI: 75.86%-81.37%) B-ACC in SIL grading (Task Tissue-Net) and 93.38% (95% CI: 88.93%-97.83%) in Silva subtyping (ADC-Silva), outperforming the baseline (UNI) by 2.00 pp and 8.96 pp ($P = 0.0022$), respectively. Notably, CerS-Path was validated on an additional public dataset (SIL-NIS), where it achieved a B-ACC of 80.04% (95% CI: 77.33%-82.74%), outperforming the baseline (UNI) by 2.37 pp, further bolstering its reliability and confidence within the open-source community. We further supplemented the dataset, improving the system's performance on cervical squamous intraepithelial lesions (SIL-NB), with a B-ACC of 94.20% (95% CI: 93.08%-95.32%). We then evaluated the SCC grading performance of CerS-Path, where the system achieved 80.63% (95% CI:

74.24%-87.03%) B-ACC, with 2.93 pp over the UNI and 7.35 pp over other comparison models. Furthermore, CerS-Path demonstrated higher diagnostic consistency than the participating pathologists in ambiguous subtyping cases, with detailed consistency presented in Supplementary Fig. 4.

Histological and tissue-type classification of cervical lesions provides therapeutic guidance[35, 36, 37, 38, 39]. We further evaluated the model's performance in subtyping tasks at both the WSI-level and the ROI-level. CerS-Path outperformed UNI on benign lesion classification (Benign-Subtype, B-ACC 81.61% (95% CI: 81.18%-83.04%)) and HPV-based adenocarcinoma subtyping (ADC-HPV, B-ACC 98.95% (95% CI: 98.19%-99.72%)) by 3.05 pp and 1.23 pp, respectively. At the ROI level, the model achieved 66.19% (95% CI: 63.64%-68.74%) B-ACC in CIN grading (ROI-CIN-9), surpassing UNI by 2.47% (Fig. 4h). In pan-tissue recognition (ROI-Pan-28), the model reached 92.36% (95% CI: 91.59%-93.14%) B-ACC, with the classification performance for each category illustrated in Fig. 4i.

Quantitative analysis of cervical tumors, such as measuring maximal diameter and invasion depth, is essential for treatment assessment and prognosis, underscoring the value of ROI segmentation[40, 41, 42]. We evaluated CerS-Path on two segmentation tasks: intraepithelial lesion segmentation (ROI-Seg-Lesion) and tumor region segmentation (ROI-Seg-Tumor), achieving an IOU of 94.64% and 86.72%, respectively. Visual results for tumor segmentation are shown in Fig. 4j, providing a reliable basis for precise quantitative computation.

Overall, with further enhancement through subspecialty-specific knowledge, it outperformed general models in screening, subtyping, grading, and ROI-level analysis, demonstrating improved baseline performance in cervical subspecialty diagnostic pathology.

## Performance of CerS-Path on Specialized Tasks

We further retrospectively evaluated the performance of the CerS-Path system on four specialized diagnostic functions: anomaly detection, rare cancer subtyping, prediction (biomarker and prognostic), and image-text multimodal analysis. Twelve tasks were assessed using downstream datasets from CerS-140K. For the subspecialty tasks, the system maintained consistent superiority in complex cervical histopathology (Fig. 5a, Supplementary Table 1), surpassing the SOTA models by an average margin of 4.34 pp in these clinically challenging scenarios.

Anomaly detection is critical as it underscores the clinical importance of detecting rare and out-of-distribution (OOD) lesions[43]. In repeated OOD detection trials, the average confidence gap—defined as the absolute difference in predicted probabilities between SCC ($p0$) and ADC ($p1$)—distinguished in-distribution (ID) samples from rare OOD entities, with a confidence interval width of 22.34% (Supplementary Table 7). Using a bimodal confidence distribution, the optimal threshold for anomaly detection yielded a detection rate of 77.71%, which exceeded UNI by 21.88 pp and the Virchow2 by 16.25 pp, while maintaining the original cancer anomaly detection task (Cancer-ADM) with a B-ACC of 98.33% (95% CI: 97.91%-98.75%). As shown in the dimension-reduced feature map (Fig. 5b, Supplementary Table 8), CerS-Path (silhouette coefficient of 0.3929) effectively distinguished anomalous categories from known classes, with a silhouette coefficient relative improvement of 8.44% over UNI (silhouette coefficient of 0.3623). Comparative results against other foundation models further substantiate the superior out-of-distribution performance of CerS-Path in uncovering rare diseases within this subspecialty (Supplementary Fig. 5).

In the rare cancer subtyping task, tumor subtype distributions are inherently long-tailed, including in cervical cancer[44, 45, 46]. We also evaluated gastric-type adenocarcinoma (ADC-Rare), a low-prevalence and aggressive subtype. CerS-Path (98.57%, 95% CI: 97.91%-98.75%) achieved a 4.05 pp increase in B-ACC over the UNI (94.52%, 95% CI: 90.29%-98.76%) and outperformed other SOTA models by an average of 7.15 pp. CerS-Path achieved superior clustering (DBI: 0.4423; silhouette coefficient: 0.4983) compared to UNI (DBI: 0.9721), and led in 5 of 6 tail-class metrics, including the highest mean cosine distance (0.4755) and strong Wasserstein distance (0.1163). These gains demonstrate stable and reliable performance, highlighting the system's robustness in detecting rare cervical adenocarcinoma. For details, see Supplementary Table 6.

Molecular expression patterns and subtypes of cervical cancer are closely associated with prognosis[47, 48, 49]. In biomarker prediction tasks, we evaluated the expression status of P16 protein (P16-pred) and MUC16 molecule (MUC16-pred). CerS-Path achieved a B-ACC improvement of 5.59 pp over UNI on the MUC16 task, with a 2.61 pp average performance advantage across models, as the model focused on key tumor regions, whereas general models showed dispersed attention (Fig. 5c). In the P16 task, CerS-Path outperformed UNI and Virchow2 models by 4.22 pp and 3.22 pp, respectively. Corresponding attention maps (Fig. 5d) confirmed spatially accurate alignment with histopathological features in P16-positive and high-risk lesion areas.

Accurate prognostic stratification is essential for managing cervical cancer[2, 50]. Using the First Affiliated Hospital of Soochow University (FHSU) and TCGA-CESC datasets, we evaluated Overall Survival (OS) prediction and Progression-Free Survival (PFS) prediction (Fig. 5e, Supplementary Fig. 6). CerS-Path achieved a global mean C-index of 0.7134 ($P$ = 0.0025) for OS prediction on the FHSU-Survival dataset, surpassing the UNI baseline (0.6515) and other prominent foundation models such as H-optimus0 and Virchow2. Similarly, for PFS on the FHSU-Progress cohort, the system yielded a C-index of 0.6975 ($P$ =

0.000056), consistently outperforming all comparison models. These results were further corroborated on the TCGA-CESC cohort (TCGA-Survival, TCGA-Progress)—which primarily comprises mid-to-late stage cervical cancer—where CerS-Path achieved an OS C-index of 0.6341($P$ =0.00060) and a PFS C-index of 0.6447 ($P$ =0.00021). Across these diverse datasets, CerS-Path demonstrated superior performance over existing state-of-the-art (SOTA) benchmarks, highlighting its capability in prognostic risk stratification.

Multimodal integration addresses complex clinical diagnostic challenges by enabling comprehensive data synthesis[49, 51, 52]. For the multimodal pathology tasks, CerS-Path gained the ability to generate region-level descriptions of cervical histopathology through feature mapping and chained integration with a large language model (LLM). As illustrated in Fig. 5f, the generated captions align well with those written by expert pathologists. Quantitatively, CerS-Path achieved a ROUGE-L of 19.72 and a BLEU-3 of 20.08, successfully outperforming state-of-the-art pathology VLMs such as Pathgen-llava[53] and Patho-R1[54]. Furthermore, it consistently exceeded the capabilities of alternative foundation models (e.g., UNI and Virchow2) evaluated within the same fine-tuning framework (Fig. 5g). Additional validation cases are provided in Supplementary Fig. 7. These results demonstrate CerS-Path's improved capacity for lesion description and its potential for handling complex clinical scenarios. Building upon its text-prompt question answering and description capabilities, the system functions as a foundational assistant for pathology report generation, enabling real-time collaborative diagnosis (Fig. 5h). By integrating a specialized subspecialty pathology knowledge base and toolkit, the simple agent system we developed (Supplementary Movie 1) demonstrates the potential to build subspecialty agents that extend system capabilities.

To assess multimodal alignment, we compared CerS-Path against a variant using random LoRA embeddings instead of CLIP-based alignment. Under identical training setups, the aligned model achieved significantly superior captioning performance. Furthermore, omitting instruction tuning caused the system to generate incoherent or non-terminating text, rendering image captioning infeasible (Supplementary Table 9). These results confirm that multimodal alignment enhances the functional utility of the vision foundation model as a multimodal embedding.

In summary, CerS-Path proved capable of handling complex, specialized diagnostic tasks. It demonstrated functionalities beyond those of general pathology models, including anomaly detection, long-tail classification, biomarker prediction, and multimodal Q&A. These capabilities highlight the system's subspecialty-level expertise and lay the foundation for its application in complex clinical diagnostic scenarios.

## Multicenter Clinical Validation of the CerS-Path System

For further prospective clinical validation, we integrated multiple diagnostic functions and deployed a digital pathology intelligence platform, as shown in Fig. 6a. The CerS-Path system also demonstrated initial capability for generating structured clinical diagnostic reports. To bridge the gap between initial benchmark results and the rigorous clinical-grade performance required for lesion and precancer screening, we first supplemented the fine-tune cohort with over 12,000 additional WSIs. Then, keeping the pretrained image encoder frozen, we trained a separate aggregator module to combine the features extracted from different patches of the whole-slide image.

To assess the clinical robustness of CerS-Path, we conducted a multicenter external validation from August 2024 to May 2025 across 5 clinical institutions and 1 public cohort, with various staining protocols, scanner types, and patient demographics. The evaluation focused on core diagnostic functions: lesion screening, precancer screening, cancer subtyping, and Silva pattern classification for adenocarcinoma, with detailed case and slide distributions per cohort provided in Fig. 6b. Detailed information on staining protocols, scanner types, and patient demographics is provided in Supplementary Tables 10 and 11.

In the prospective lesion screening evaluation, which estimates a model's ability to detect any pathological lesion (LSIL, HSIL, and Cancer) against negative/benign backgrounds, clinical testing of CerS-Path was conducted across an expanded network, including Ningbo Pathology Center (NBPC), Jinhua Central Hospital (JHC), Guangdong Women and Children Hospital (GWCH), ADICON, and Tissue-Net. As shown in Fig. 6c, CerS-Path consistently achieved high sensitivities exceeding 99.0% across all cohorts while maintaining high specificities (above 84.4%) in these multi-institutional test sets. The CerS-Path system demonstrated high generalizability and reliability across cohorts, with average AUC scores of 98.61% (peaking at 99.73%) and outperformed leading foundation models (UNI, Virchow2, and H-optimus0).

In the prospective precancer screening evaluation, which measures the accurate identification of high-risk actionable lesions (HSIL and invasive carcinomas) versus non-actionable cases (Benign and LSIL), clinical testing of CerS-Path was conducted across an expanded network including NBPC, JHC, GWCH, ADICON, and Tissue-Net. As shown in Fig. 6d, CerS-Path consistently achieved sensitivities exceeding 98.6% across all prospective cohorts, while maintaining high specificities (above 88.4%) in all multi-institutional test sets. Furthermore, the system maintained robust AUC scores above 98.1% (peaking at 99.34%)—outperforming foundation models like UNI and Virchow2 in four of the five cohorts, confirming its highly generalizable reliability for clinical deployment in early precancer screening.

In the cancer subtyping evaluation, which refers to the diagnosis and differentiation of the relatively rare and morphologically challenging ADC from the more common SCC, clinical testing of CerS-Path was conducted across a multi-center network including NBPC, GWCH, and XGCH. As shown in Fig. 6e, CerS-Path consistently achieved sensitivity exceeding 98.9% across all prospective cohorts, outperforming leading foundation models, while maintaining high specificities (above 85.0%) in all multi-institutional test sets. The robust AUC performance further confirmed its highly generalizable reliability for clinical deployment in precise tumor subtyping.

CerS-Path also addressed the highly challenging Silva classification task. In the XGCH cohort, as shown in Fig. 6f, the system achieved an F1-score of 80.87% and a B-ACC of 78.93%, indicating strong consistency.

In summary, CerS-Path demonstrated robust diagnostic performance across the full spectrum of cervical histopathology, while the subspecialty enhancement strategy offers avenues for translational research in computational pathology.

## Discussion

Existing pathology foundation models exhibit critical limitations in cervical cancer diagnosis due to the histopathological complexity and clinical exigencies. In this study, we bridged this gap by introducing CerS-Path, developed via self-supervised learning on large-scale curated cervical histopathology slides and enhanced with multimodal capabilities. This yields a vertically specialized diagnostic system that advances pathology foundation models through cervical specialization while expanding their clinical application scope.

A core advantage of CerS-Path is its capacity in precision subspecialty diagnosis. It elevates professional-level clinical utility by establishing state-of-the-art (SOTA) benchmarks in 26 out of 30 tasks. These tasks span a comprehensive suite of eight diagnostic functions, including screening, grading, subtyping, quantitative analysis, rare cancer detection, predictive modeling, multimodal Q&A, and automated reporting. This work represents progress in adapting general-purpose foundation models to subspecialty-specific diagnostic tasks in computational pathology.

CerS-Path highlights the value of vertical specialization by outperforming the evaluated general-purpose models on subspecialty-specific tasks. The injection of cervical subspecialty priors enables the precise decoding of fine-grained morphological semantics, driving substantial improvements in challenging tasks like Adenocarcinoma Silva pattern classification. Notably, in low-consensus domains such as SCC grading, the model achieves consistency comparable to expert pathologists. Additionally, CerS-Path establishes a robust diagnostic safety net through out-of-distribution (OOD) detection of rare carcinomas, a function inherently absent in generalist baselines. By outperforming massive-scale models in advanced tasks—including long-tailed rare malignancy identification, biomarker prediction, survival prognostication, zero-shot classification, and multimodal captioning—CerS-Path proves that domain-specific priors are essential for achieving comprehensive diagnostic reliability in clinical practice.

Furthermore, the specialization strategy of CerS-Path offers a distinct pathway to high-performance clinical diagnosis beyond the mere scaling of data and computation. By enhancing various pathology foundation models with different architectures and initializations (e.g., CTransPath[55], Phikon[56]), we validated the effectiveness of this approach in both evaluated models (Supplementary Table 12). Notably, subspecialty-enhanced lightweight models achieved performance parity with, or even superiority over, massive general foundation models, despite a 40-fold difference in parameter scale (Supplementary Table 13). As computational pathology moves toward large-scale clinical integration, this specialized efficiency will substantially reduce the computational overhead required for high-volume diagnostic workflow.

In prospective and retrospective evaluations across 5 clinical institutions and 1 public cohort, CerS-Path demonstrated consistent performance, maintaining high sensitivity in core tasks including Lesion Screen, Precancer Screen, and Cancer Subtype. Despite the heterogeneity of staining protocols, scanner hardware, and patient demographics, its comprehensive performance outperformed generalist foundation models (Supplementary Fig. 8, Supplementary Table 11). Crucially, CerS-Path maintained high sensitivity across all multi-center comparisons, thereby helping to minimize false-negative risks. While its specificity exhibited slight fluctuations, this reflects a deliberate clinical trade-off. To prioritize sensitivity during real-world deployment, the system adopts conservative, high-sensitivity thresholds; the resulting manageable increase in false positives represents a trade-off intended to reduce false-negative risk. Furthermore, although real-world domain shifts induced by hardware and staining variances affected all evaluated models (Supplementary Fig. 8), CerS-Path still achieved a higher average specificity than generalist baselines across all scenarios. Consequently, by maintaining conservative diagnostic thresholds while effectively reducing the overall review burden for pathologists, CerS-Path demonstrates potential for clinical translation.

While achieving over 90% accuracy in core functions, the system requires further refinement in high-level predictive and prognostic tasks. Notably, although these predictive tasks currently fall short of clinical-grade proficiency, they provide valuable insights by identifying morphological features that are not directly perceptible to the human eye. For challenging scenarios with low consensus, such as SCC grading, CerS-Path serves as a diagnostic aid to enhance workflow efficiency and consistency. For CerS-Path, tasks with sub-90% accuracy either address inherently difficult morphological classifications where the model's role is to support rather than replace human judgment, or involve advanced predictive or out-of-distribution tasks that provide clinically valuable triage and reference information. In these cases, the nature of model errors is such that they can be identified and corrected by pathologists, while the overall system enhances diagnostic efficiency and safety. For general morphological tasks such as lymph node metastasis detection (LNM), where the morphological features of cancer infiltration are highly conserved across different cancer types, current state-of-the-art foundation models such as UNI may already encode these general representations during pan-cancer pre-training. We suggest that such capabilities may generalize across organ-agnostic foundations. However, for the sake of diagnostic comprehensiveness, these functions remain integrated within CerS-Path to support a unified clinical workflow.

Future research will enhance CerS-Path's capabilities across four critical dimensions. First, improving adaptability to dynamic resolution remains essential. Current diagnostic tasks rely on patch-level analysis, limiting adaptation to variable spatial granularity requirements. This constraint impedes effectiveness in resolution-sensitive scenarios such as microinvasion detection in cervical squamous cell carcinoma and complex multimodal visual question answering. Multi-scale modeling strategies will be explored to extend the receptive field and achieve resolution-adaptive performance. Besides, the current implementation supports only basic structured output and limited contextual interaction. Sophisticated logical reasoning and deeper clinical integration necessitate developing a cervical pathology-specific planning engine, enabling automated diagnostic logic and seamless workflow integration.

Furthermore, algorithmic efficiency and embedded deployment must be prioritized to reduce interaction costs and enhance usability. These improvements will facilitate real-time applications including disease progression analysis, therapeutic monitoring, and morpho-molecular biomarker discovery. Finally, further validation through expanded multi-center trials across multiple diagnostic functionalities will assess structured report reliability and clinical readiness.

In summary, CerS-Path is a representative system for cervical subspecialty pathology, advancing foundation models toward specialized diagnosis. By supporting full clinical integration, CerS-Path expands the role of foundation models from general representation to system-level deployment. This represents a step from organ-agnostic representation models toward deployable, workflow-aligned subspecialty tools for cervical histopathology, supporting further clinical evaluation.

# Methods

## Ethical approval

This study complied with all relevant ethical regulations and the Declaration of Helsinki. The use of de-identified pathological images and associated data was approved by the Ethics Committee of the Ningbo Clinical Pathology Diagnosis Center, Ningbo, China (approval no. BLZX20250701); the Medical Ethics Review Committee of Jinhua Central Hospital, Jinhua, China (approval no. (Research) 2026-Ethics Review-428); the Medical Ethics Committee of the First Affiliated Hospital of Soochow University, Suzhou, China (approval no. [2025] 789); the Hangzhou ADICON Clinical Laboratory Ethics Committee,

Hangzhou, China (approval no. ADICON [2026] 001); the Medical Ethics Committee of Xiaogan Central Hospital, Xiaogan, China (approval no. KY-2025063007); and the Medical Ethics Committee of Guangdong Women and Children Hospital, Guangzhou, China (approval no. 20260001). The requirement for written informed consent was waived by the respective ethics committees because the study used previously collected clinical data and specimens, protected participant identity and privacy, and required no further contact or follow-up. Publicly available de-identified datasets did not require additional ethical approval. Participant sex was based on sex assigned at birth, as documented in electronic medical records for institutional cohorts and in the accompanying metadata for public cohorts; all participants were recorded as female.

## Construction of CerS-140K: A Pretrained Cervical Subspecialty Pathology Dataset

General-purpose pathological foundation models exhibit a lack of precise subspecialty representation and insufficient comprehensive benchmarks for clinical diagnosis. To address these dual limitations, we constructed CerS-140K, a large-scale dataset of ~140,000 cervical whole-slide images (WSIs). The in-house component comprises biopsies and excisions collected from the Ningbo Pathology Center (2021–2024) and digitized using an SQS-600 scanner (Supplementary Fig. 9a). To strictly isolate training from evaluation, CerS-140K was partitioned into two functional subsets. Specifically, ~115,000 purely unannotated WSIs, categorized solely by disease type, were dedicated to the self-supervised pretraining stage (Stage I in Fig. 2a), with the disease types distribution shown in Fig. 1a. Within this subset, primary tumor cases (e.g., ADC and SCC) contribute an average of 5–10 slides per case, ensuring substantial morphological diversity. In contrast, the remaining ~25,000 WSIs—integrating in-house cohorts with public benchmarks—were rigorously expert-annotated for downstream evaluation and clinical enhancement (Stage IV in Fig. 2c). The detailed distribution and characteristics of CerS-140K are presented in Supplementary Table 14.

WSIs were preprocessed using a customized tissue segmentation pipeline designed for histopathology. Initial segmentation was performed on down-sampled 1.25× resolution images in the RGB color space, followed by boundary refinement through median blurring, morphological closing, and artifact filtration to produce smooth tissue contours. Final patch extraction was performed on 20× magnification WSIs, with non-overlapping 256×256 regions extracted from segmented tissue areas.

To ensure adequate coverage for pretraining, we curated about 80,000 biopsy slides from an initial pool of 115,000 cervical biopsies and extracted approximately one-third of the patches from each WSI, yielding ~100 million non-overlapped 256×256-pixel patches at 20× magnification. An additional 20,000 conization or resection specimens yielded 80 million patches through 50% tissue area sampling per slide. Finally, 15,000 WSIs derived from lymph nodes, the lower cervix, or other pelvic sites contributed 10 million patches, with 0.5-1K patches per slide. This hierarchical curation strategy produced 190 million diagnostically relevant patches for specialized model pretraining.

To mitigate inter-observer variability, ground truth labels were established through a rigorous consensus protocol where all cases underwent independent blind review by at least two senior subspecialty pathologists. Discordant cases were escalated to an arbitration committee of chief pathologists with at least 15 years of experience for final consensus. Crucially, diagnoses for morphologically equivocal cases were definitively anchored using objective clinical and molecular data—including a comprehensive panel of immunohistochemical markers, molecular genotyping, and subsequent surgical excision reports—ensuring robust and unbiased evaluation benchmarks.

## Visual Representation Enhancement Pretraining Framework

General-purpose pretrained models often struggle to adapt to pathology subspecialties, particularly in domains with scarce training data[55], compounded by the fragmented and insufficient availability of public datasets for pathology subspecialties. Thus, we conducted large-scale pretraining of CerS-Path on our curated cervical pathology dataset, CerS-140K.

The overall visual pretraining framework is illustrated in Fig. 2a, Stage I. We employed the SOTA DINOv2 architecture [28], a self-supervised vision training framework based on a teacher–student paradigm. Within this paradigm, the student model learned to align its output with pseudo-labels generated by the teacher model, enabling knowledge distillation through feature distribution matching. The core strength of DINOV2 lies in its sophisticated contrastive learning mechanism, where positive and negative samples within each batch were processed through dual teacher-student pathways to compute reconstruction and alignment losses.

For each batch of cervical tissue image patches, random augmentations, including color jitter, Gaussian blur, and exposure adjustment, were applied to generate multiple augmented views, and produced multiple augmented views consisting of two global crops alongside several local crops. A masking strategy was applied to one of the global crops to support masked image modeling. The unmasked global crop served as inputs to the teacher model, while masked global crops and local crops were fed into the student model to produce patch-level prototypes. The training optimization involved two key components: a reconstruction objective that trained the student model to predict teacher outputs in masked regions using surrounding contextual information, and an alignment objective that reduced the cross-entropy between aggregated student representations (derived from both masked global and local crops) and their corresponding teacher representations from unmasked views. The teacher model was updated as an exponential moving average of the student parameters.

During training, we initialized the model with weights from UNI and set the patch size to 16. The training patch covered an area of 20× on 256×256 images. The learning rate was set to 0.0002, and the local-crop-size parameter was configured as 96. All experiments were conducted on a single A100 node using Fully Sharded Data Parallel (FSDP) for distributed training.

## Multimodal Representation Enhancement

Multimodal pretraining has recently been introduced as an auxiliary strategy to augment pathology foundation models, equipping them to address increasingly complex and prevalent clinical scenarios[57, 58, 59, 60, 61]. However, the scarcity of high-quality paired pathology data, particularly in subspecialty domains, and the predominance of alignment-based multimodal pretraining approaches have restricted the development of specialized tissue representations[24].

To support multimodal integration, we constructed a dataset of approximately 2.5 million patch-level image-text pairs (Supplementary Table 15). This dataset integrates large-scale public pathology libraries (PathGen[53], Quilt-1M[62], and PathCap[63]) with an in-house authoritative dataset extracted from digitized medical textbooks using DocLayout-YOLO[64]. Following data curation to standardize text length and filter semantic content, the refined dataset was utilized for multimodal alignment pretraining. We augmented our visual feature extractor (CerS-V) with an auxiliary low-rank adaptation (LoRA) module [65] to enable multimodal learning (Fig. 2b). This approach preserves CerS-V's robust visual representational capacity for cervical histopathology while extending its capabilities to process textual inputs, yielding the multimodal embedding model CerS-M. CerS-M involves three key components: a visual encoder (CerS-V) augmented with a LoRA module, a mapping layer implemented as a multi-layer perceptron (MLP), and a large language model (LLM), for which we employed Qwen-VL 2.5 (7B).

The multimodal pretraining proceeds in two stages. During the cross-modal representation alignment stage (Fig. 2b, Stage II), we aligned the LoRA-augmented visual encoder and the MLP projection layer using image-text pairs collected from both public and private sources. Inspired by Coca [66], this step jointly optimized the multimodal representation capacity of the visual encoder and its interface with the LLM. The details of the multimodal extension model are shown in Supplementary Fig. 9b.

In detail, instead of updating the full-rank projection matrix $\mathbf{W}_0 \in \mathbb{R}^{d_2 \times d_1}$, we introduce a learnable LoRA module as:

$$\widetilde{\mathbf{W}} = \mathbf{W}_0 + \alpha \cdot \mathbf{UV}, \tag{1}$$

where $\mathbf{U} \in \mathbb{R}^{d_2 \times r}, \mathbf{V} \in \mathbb{R}^{r \times d_1}$, and $r \ll \min(d_1, d_2)$. During training, $\mathbf{U}$ and $\mathbf{V}$ are trainable, while $\mathbf{W}_0$ remains frozen. Following standard practice, we use Gaussian initialization for $\mathbf{V}$ and zero initialization for $\mathbf{U}$, ensuring $\widetilde{\mathbf{W}} = \mathbf{W}_0$ at the start of training. This formulation enables efficient fine-tuning with minimal additional parameters and maintains model stability.

The second stage (Stage III in Fig. 2b) involved instruction tuning of the LLM for domain-specific pathology question-answering tasks. To enable instruction tuning for downstream multimodal tasks, we further curated a specialized dataset by extracting cervical tissue-relevant samples from publicly available resources. We selected 3,890 high-quality instances and reformulated them into instruction-style prompts using GPT-4o, adhering to the ShareGPT dialogue schema. In this reformulation, captions were expanded into natural language instructions to simulate realistic human-AI interactions. During fine-tuning, we froze CerS-V's vision backbone to preserve pretrained representations while making the LoRA modules, multimodal alignment MLP, and LLM components tunable. The rank r of LoRA module was set to 64, with $\alpha = 64$ and a dropout of 0.25. The MLP projection dimension is set to 3584. Flash attention was enabled, and training was performed with dynamic learning rate scheduling.

## Evaluation Setting

CerS-Path demonstrated broad applicability across diverse downstream tasks (Fig. 2c). This section details our evaluation strategies covering four dimensions: WSI-level tasks, ROI-level analysis, multimodal integration, and multi-cohort generalization. All weakly supervised classification tasks in this study, including screening, lesion grading, WSI-level few-shot learning, long-tailed dataset classification, and predictive modeling, were implemented using the CLAM framework [67]. The following section details the corresponding evaluation criteria and procedures. Details of the downstream fine-tuning and evaluation datasets are provided in Supplementary Note. Fold-level classification metrics were compared using one-sided paired t-tests across five matched cross-validation folds, testing whether CerS-Path outperformed each comparator. Survival distributions were compared using two-sided log-rank tests. $P < 0.05$ was considered statistically significant, with no adjustment for multiple comparisons.

## Evaluation on WSI-Level Tasks

For WSI-level tasks, we followed the WSI preprocessing pipeline detailed in the dataset construction section. Tissue segmentation was performed using a U-Net model with a ResNet-34 based encoder (detailed in Supplementary Note 1), trained on low-resolution histopathological thumbnails. Pretrained encoders were applied at the patch coordinate level to extract feature representations.

WSI patches were sampled at 20× magnification, with an initial patch size of 256×256 pixels. Prior to embedding, patches were resized to 224×224 pixels and normalized using ImageNet mean and standard deviation values for feature consistency.

We employed the two-branch gated variant of the CLAM architecture, where the initial fully connected layer projects all input embeddings into a 512-dimensional latent space, followed by an intermediate hidden layer of 384 dimensions. A dropout rate of 0.25 was applied after each intermediate layer to improve regularization.

For anomaly detection, we extended the original CLAM training objective by incorporating an open-set detection strategy based on the ARPL loss [31]. This approach introduces reciprocal points to construct an auxiliary class space for known categories, thereby reducing the overlap between known and unknown distributions. To quantify diagnostic certainty and identify out-of-distribution (OOD) anomalies, we defined a confidence score based on the predictive margin:

$$\Delta p = |p_1 - p_0| \quad (2)$$

where $p_1$ and $p_0$ denote the activation probabilities for adenocarcinoma and squamous cell carcinoma, respectively. For typical in-distribution samples, an ideal model yields a definitive diagnosis with a large $\Delta p$ (approaching 1). Conversely, for morphologically atypical OOD samples, the model reflects diagnostic uncertainty, yielding a significantly smaller $\Delta p$. General-purpose deep learning models frequently exhibit overconfidence bias on unfamiliar inputs. By penalizing overconfident activations for OOD samples, this mechanism enhances the model's ability to reject unknown classes while preserving classification accuracy on in-distribution categories.

Prognostic evaluation utilized five-fold cross-validation and Attention-Based Multiple Instance Learning (ABMIL) for feature aggregation[68]. The model minimized Cox[69] loss via the Adam optimizer (learning rate = 2e-4, weight decay = 1e-5). After excluding cases with insufficient follow-up, we evaluated OS and PFS across the early-stage FHSU cohort (TI–TII) and the advanced-stage TCGA-CESC dataset (TII–TIII+). Patients were stratified by the median risk score, with prognostic performance assessed using the C-index and Kaplan-Meier survival curves (log-rank test).

## ROI-Level Evaluation

For ROI-level classification tasks, we adopted a linear probe strategy to evaluate the discriminative power of pretrained feature embeddings in downstream tasks. The input images were resized to 224×224 pixels and normalized using ImageNet mean and standard deviation parameters. This fixed-dimension preprocessing accommodates Virchow2's architectural constraints while maintaining compatibility across models, despite many ROI datasets containing higher-resolution source images.

During linear probe evaluation, the weights of the pretrained pathology backbone models (UNI, Virchow2, ConchV15 and CerS-Path) were kept frozen. All linear classifiers were trained with the Adam optimizer and ReduceLROnPlateau learning rate scheduler, with an initial learning rate of $1\times10^{-4}$. $\ell_2$-regularization was applied with the coefficient $\lambda = \frac{100}{MC}$, where $M$ denotes the embedding dimension and $C$ represents the number of target classes. Cross-entropy loss was used as the optimization objective, and training was conducted for a maximum of 80 epochs.

For ROI-level segmentation tasks, we extracted 1024×1024-pixel patches from WSIs at 20× magnification and resized them to 224×224 pixels, applying ImageNet-style normalization. Mirroring the classification approach, linear segmentation heads were optimized using Adam with ReduceLROnPlateau (initial learning rate: $1\times10^{-4}$), Dice loss, and an 80-epoch training regimen.

## Multimodal Pathology Integration Evaluation

The multimodal capabilities of CerS-Path were evaluated through two complementary strategies: pathology caption generation and text-guided zero-shot classification.

For caption generation, we designed a multimodal prompt-response framework. For each ROI image, a standardized prompt "Please diagnose the cervical pathology image.", was provided as input. Model-generated diagnostic descriptions were quantitatively evaluated against reference captions authored by board-certified cervical histopathology experts using five automated metrics: ROUGE-L, BLEU-1, BLEU-3, BLEU-5, and semantic similarity (Bert, GPT). Reference captions were written by board-certified pathologists with expertise in cervical histopathology and served as the ground truth. All metrics were computed using open-source Python implementations, without manual intervention or post-processing. This evaluation framework enables a direct assessment of the alignment between model-generated and expert-authored diagnostic descriptions, and reflects the model’s capacity for multimodal understanding and expression in cervical pathology.

For zero-shot classification, we designed a prompt-guided inference paradigm. For non-generative baselines (e.g., CerS-M, CONCH, UNI), we followed the contrastive zero-shot protocol. Specifically, we computed the dot product between the L2-normalized image features and text features derived from class-specific prompts (e.g., "an H&E image of Benign lesion"), with detailed categorization (benign vs. malignant) criteria presented in Supplementary Table 16. Final predictions were then determined by the highest softmax probability across all candidate classes. For the LLM-integrated CerS-Path, we utilized a prompt-guided inference paradigm where images were queried with a standardized prompt ("Please diagnose the cervical pathology image. Is it cancerous? Answer 'Yes' or 'No'."), and the generated text responses were directly mapped to binary diagnostic labels.

All predictions were compared against ground-truth labels using five complementary metrics: Accuracy, Balanced Accuracy (addressing class imbalance), F1-score, Precision, and Recall. Evaluations strictly prohibited dialogue history, contextual memory, or additional fine-tuning to objectively assess real-world generalization under clinical constraints.

## Multi-Cohort Generalization Evaluation

To evaluate the clinical stability and generalizability of the CerS-Path subspecialty diagnostic system, we conducted a comprehensive multi-institutional validation. This evaluation encompassed both prospective and retrospective cohorts from diverse sources, including Ningbo Pathology Center (NBPC, prospective), Guangdong Women and Children Hospital (GWCH, prospective), Jinhua Central Hospital (JHC, prospective), Hangzhou Adicon (ADICON, prospective), Xiaogan Central Hospital (XGCH, mixed), and the French Tissue-Net (retrospective). The assessment encompassed key diagnostic workflows including cervical cancer screening, histological subtyping, and adenocarcinoma Silva pattern classification.

Within an integrated software-clinical deployment framework, board-certified pathologists interacted with dedicated diagnostic modules. For classification tasks within the clinical setting, we analyzed the weakly supervised predictions generated by CerS-Path and computed standard diagnostic performance metrics, including the area under the receiver operating characteristic curve (AUC), sensitivity, and specificity. Sensitivity and specificity are defined mathematically as follows:

$$\text{Sensitivity} = \frac{\text{TP}}{\text{TP} + \text{FN}}, \tag{3}$$

$$\text{Specificity} = \frac{\text{TN}}{\text{TN} + \text{FP}}, \tag{4}$$

where $\text{TP}, \text{TN}, \text{FP},$ and $\text{FN}$ denote the number of true positives, true negatives, false positives, and false negatives, respectively.

For binary classification tasks such as cervical disease screening and cancer subtyping, we calibrated activation thresholds to prioritize detection of positive cases (e.g., adenocarcinoma), targeting a threshold of 0.8 to align with clinical properties. For multiclass diagnostic applications, we maintained default thresholds and documented real-world performance using F1-scores and balanced accuracy (B-ACC).

## Data Availability

The processed evaluation data generated in this study are provided in the Supplementary Information and accompanying Source Data file. Detailed descriptions of the datasets are provided in Supplementary Note 1. The raw in-house whole-slide images and associated clinical metadata are protected and cannot be made publicly available owing to the applicable ethics approvals, patient privacy requirements and data-protection regulations. The authors are not authorized to share these data. The public data used in this study are available from their original repositories: TissueNet (https://doi.org/10.60597/eaqa-k904), SIL-NIS (https://rdm.inesctec.pt/dataset/nis-2024-003), TCGA (https://portal.gdc.cancer.gov/projects/TCGA-CESC) and ROI-CIN-9 (https://biod.whu.edu.cn/sjj.htm). Source data are provided with this paper.

## Code Availability

The source code is publicly available at GitHub (https://github.com/rainyfog/CerS-Path), and the version used in this study is permanently archived on Zenodo (https://doi.org/10.5281/zenodo.21770459). A demonstration program is available at Code Ocean (https://codeocean.com/capsule/2456104/tree). The original CerS-Path code is released under the MIT License without access restrictions. Third-party components are attributed in the repository and remain subject to their original licenses.

## References

1. Siegel RL, Giaquinto AN, Jemal A. Cancer statistics, 2024. *CA: a cancer journal for clinicians* **74**, 12–49-12–49 (2024).

2. Cohen PA, Jhingran A, Oaknin A, Denny L. Cervical cancer. *The Lancet* **393**, 169–182-169–182 (2019).

3. Ouh Y-T*, et al.* Discrepancy between cytology and histology in cervical cancer screening: a multicenter retrospective study (KGOG 1040). *Journal of Korean Medical Science* **36**, (2021).

4. Jiang P*, et al.* A systematic review of deep learning-based cervical cytology screening: from cell identification to whole slide image analysis. *Artificial Intelligence Review* **56**, 2687–2758-2687–2758 (2023).

5. Ronco G*, et al.* Efficacy of HPV-based screening for prevention of invasive cervical cancer: follow-up of four European randomised controlled trials. *The lancet* **383**, 524-532 (2014).

6. Darragh TM*, et al.* The lower anogenital squamous terminology standardization project for HPV-associated lesions: background and consensus recommendations from the College of American Pathologists and the American Society for Colposcopy and Cervical Pathology. *Archives of pathology & laboratory medicine* **136**, 1266-1297 (2012).

7. Melnikow J, Henderson JT, Burda BU, Senger CA, Durbin S, Weyrich MS. Screening for cervical cancer with high-risk human papillomavirus testing: updated evidence report and systematic review for the US Preventive Services Task Force. *Jama* **320**, 687-705 (2018).

8. Stelzle D*, et al.* Estimates of the global burden of cervical cancer associated with HIV. *The lancet global health* **9**, e161–e169-e161–e169 (2021).

9. Qu Z, Wang Z, Qiu S, Cui G, Li C. Efficacy of photodynamic therapy with 5-aminolevulinic acid for the treatment of cervical high-grade squamous intraepithelial lesions with high-risk HPV infection: A retrospective study. *Photodiagnosis and Photodynamic Therapy* **40**, 103068-103068 (2022).

10. Kojima A*, et al.* Gastric morphology and immunophenotype predict poor outcome in mucinous adenocarcinoma of the uterine cervix. *The American journal of surgical pathology* **31**, 664–672-664–672 (2007).

11. Perkins RB, Wentzensen N, Guido RS, Schiffman M. Cervical cancer screening: a review. *Jama* **330**, 547–558-547–558 (2023).

12. Khodakarami N, Farzaneh F, Aslani F, Alizadeh K. Comparison of Pap smear, visual inspection with acetic acid, and digital cervicography as cervical screening strategies. *Archives of gynecology and obstetrics* **284**, 1247–1252-1247–1252 (2011).

13. Li Y-x, Chen F, Shi J-j, Huang Y-l, Wang M. Convolutional neural networks for classifying cervical cancer types using histological images. *Journal of Digital Imaging* **36**, 441–449-441–449 (2023).

14. Meng Z, Zhao Z, Li B, Su F, Guo L. A cervical histopathology dataset for computer aided diagnosis of precancerous lesions. *IEEE Transactions on Medical Imaging* **40**, 1531–1541-1531–1541 (2021).

15. Cho B-J*, et al.* Automated diagnosis of cervical intraepithelial neoplasia in histology images via deep learning. *Diagnostics* **12**, 548-548 (2022).

16. Wang R*, et al.* A Deep Learning Framework for Predicting Prognostically Relevant Consensus Molecular Subtypes in HPV-Positive Cervical Squamous Cell Carcinoma from Routine Histology Images. *bioRxiv*, 2024–2008-2024–2008 (2024).

17. Liu Q*, et al.* A histopathologic image analysis for the classification of endocervical adenocarcinoma silva patterns depend on weakly supervised deep learning. *The American Journal of Pathology* **194**, 735–746-735–746 (2024).

18. Guo Q*, et al.* Predicting lymph node metastasis from primary cervical squamous cell carcinoma based on deep learning in histopathologic images. *Modern Pathology* **36**, 100316-100316 (2023).

19. Wang R*, et al.* Deep learning for predicting prognostic consensus molecular subtypes in cervical cancer from histology images. *npj Precision Oncology* **9**, 11-11 (2025).

20. Chen RJ*, et al.* Towards a general-purpose foundation model for computational pathology. *Nature Medicine* **30**, 850–862-850–862 (2024).

21. Xu H, *et al.* A whole-slide foundation model for digital pathology from real-world data. *Nature* **630**, 181–188-181–188 (2024).

22. Vorontsov E, *et al.* Virchow: A million-slide digital pathology foundation model. *arXiv preprint arXiv:230907778*, (2023).

23. Wang X, *et al.* A pathology foundation model for cancer diagnosis and prognosis prediction. *Nature* **634**, 970–978-970–978 (2024).

24. Ma J, *et al.* PathBench: A comprehensive comparison benchmark for pathology foundation models towards precision oncology.) (2025).

25. Campanella G, *et al.* A clinical benchmark of public self-supervised pathology foundation models. *Nature Communications* **16**, 3640 (2025).

26. Ferber D, *et al.* Development and validation of an autonomous artificial intelligence agent for clinical decision-making in oncology. *Nature cancer*, 1-13 (2025).

27. Bai S, *et al.* Qwen2. 5-vl technical report. *arXiv preprint arXiv:250213923*, (2025).

28. Oquab M, *et al.* Dinov2: Learning robust visual features without supervision. *arXiv preprint arXiv:230407193*, (2023).

29. Radford A, *et al.* Learning Transferable Visual Models From Natural Language Supervision. In: *Proceedings of the 38th International Conference on Machine Learning* (eds Meila M, Zhang T). PMLR.

30. Zimmermann E, *et al.* Virchow2: Scaling self-supervised mixed magnification models in pathology. *arXiv preprint arXiv:240800738*, (2024).

31. Chen G, Peng P, Wang X, Tian Y. Adversarial reciprocal points learning for open set recognition. *IEEE Transactions on Pattern Analysis and Machine Intelligence* **44**, 8065–8081-8065–8081 (2021).

32. Peto J, Gilham C, Fletcher O, Matthews FE. The cervical cancer epidemic that screening has prevented in the UK. *The Lancet* **364**, 249-256 (2004).

33. Holl K, *et al.* Human papillomavirus prevalence and type-distribution in cervical glandular neoplasias: Results from a E uropean multinational epidemiological study. *International journal of cancer* **137**, 2858–2868-2858–2868 (2015).

34. Pimenta JM, Galindo C, Jenkins D, Taylor SM. Estimate of the global burden of cervical adenocarcinoma and potential impact of prophylactic human papillomavirus vaccination. *BMC cancer* **13**, 1–12-11–12 (2013).

35. Park KJ. Cervical adenocarcinoma: integration of HPV status, pattern of invasion, morphology and molecular markers into classification. *Histopathology* **76**, 112–127-112–127 (2020).

36. Das DK. Therapy-Relevant Cervical Cancer Subtypes Identified.). LWW (2023).

37. Mabuchi S, *et al.* Population-based survival analysis of stage IVB small-cell neuroendocrine carcinoma in comparison to major histological subtypes of cervical cancer. *Current Oncology* **30**, 9428–9436-9428–9436 (2023).

38. Liu P, *et al.* Comparison of survival outcomes between squamous cell carcinoma and adenocarcinoma/adenosquamous carcinoma of the cervix after radical radiotherapy and chemotherapy. *BMC cancer* **22**, 326-326 (2022).

39. Saavedra KP, Brebi PM, Roa JCS. Epigenetic alterations in preneoplastic and neoplastic lesions of the cervix. *Clinical epigenetics* **4**, 1–7-1–7 (2012).

40. Jolly S, Uppal S, Bhatla N, Johnston C, Maturen K. Improving global outcomes in cervical cancer: the time has come for international federation of gynecology and obstetrics staging to formally incorporate advanced imaging. *Journal of Global Oncology* **4**, JGO–2016-JGO–2016 (2017).

41. Plante M, Gregoire J, Renaud M-C, Roy M. The vaginal radical trachelectomy: an update of a series of 125 cases and 106 pregnancies. *Gynecologic oncology* **121**, 290–297-290–297 (2011).

42. Mohamud A, Høgdall C, Schnack T. Prognostic value of the 2018 FIGO staging system for cervical cancer. *Gynecologic Oncology* **165**, 506–513-506–513 (2022).

43. Linmans J, Elfwing S, van der Laak J, Litjens G. Predictive uncertainty estimation for out-of-distribution detection in digital pathology. *Medical Image Analysis* **83**, 102655-102655 (2023).

44. Liu M, Li X, Gao X, Chen J, Shen L, Wu H. Sample Hardness Based Gradient Loss for Long-Tailed Cervical Cell Detection. In: *Medical Image Computing and Computer Assisted Intervention – MICCAI 2022* (eds Wang L, Dou Q, Fletcher PT, Speidel S, Li S). Springer Nature Switzerland.

45. Zheng Q*, et al.* Large-scale long-tailed disease diagnosis on radiology images. *Nature Communications* **15**, 10147-10147 (2024).

46. Zhang Y, Kang B, Hooi B, Yan S, Feng J. Deep long-tailed learning: A survey. *IEEE transactions on pattern analysis and machine intelligence* **45**, 10795–10816-10795–10816 (2023).

47. Gadducci A, Guerrieri ME, Greco C. Tissue biomarkers as prognostic variables of cervical cancer. *Critical reviews in oncology/hematology* **86**, 104–129-104–129 (2013).

48. Arip M, Tan LF, Jayaraj R, Abdullah M, Rajagopal M, Selvaraja M. Exploration of biomarkers for the diagnosis, treatment and prognosis of cervical cancer: a review. *Discover Oncology* **13**, 91-91 (2022).

49. Network CGAR, others. Integrated genomic and molecular characterization of cervical cancer. *Nature* **543**, 378-378 (2017).

50. Arbyn M*, et al.* Estimates of incidence and mortality of cervical cancer in 2018: a worldwide analysis. *The Lancet Global Health* **8**, e191-e203 (2020).

51. Lipkova J*, et al.* Artificial intelligence for multimodal data integration in oncology. *Cancer cell* **40**, 1095–1110-1095–1110 (2022).

52. Boehm KM, Khosravi P, Vanguri R, Gao J, Shah SP. Harnessing multimodal data integration to advance precision oncology. *Nature Reviews Cancer* **22**, 114–126-114–126 (2022).

53. Sun Y*, et al.* Pathgen-1.6 m: 1.6 million pathology image-text pairs generation through multi-agent collaboration. *arXiv preprint arXiv:240700203*, (2024).

54. Zhang W*, et al.* Patho-r1: A multimodal reinforcement learning-based pathology expert reasoner. In: *Proceedings of the AAAI Conference on Artificial Intelligence*) (2026).

55. Wang X*, et al.* Transformer-based unsupervised contrastive learning for histopathological image classification. *Medical image analysis* **81**, 102559-102559 (2022).

56. Filiot A*, et al.* Scaling self-supervised learning for histopathology with masked image modeling. *MedRxiv*, 2023.2007. 2021.23292757 (2023).

57. Xiang J*, et al.* A vision–language foundation model for precision oncology. *Nature*, 1–10-11–10 (2025).

58. Zhang L, Yang Q, Agrawal A. Assessing and Learning Alignment of Unimodal Vision and Language Models. *arXiv preprint arXiv:241204616*, (2024).

59. Li C*, et al.* Llava-med: Training a large language-and-vision assistant for biomedicine in one day. *Advances in Neural Information Processing Systems* **36**, 28541–28564-28541–28564 (2023).

60. Seyfioglu MS, Ikezogwo WO, Ghezloo F, Krishna R, Shapiro L. Quilt-LLaVA: Visual Instruction Tuning by Extracting Localized Narratives from Open-Source Histopathology Videos.) (2025).

61. Zhang K*, et al.* A generalist vision–language foundation model for diverse biomedical tasks. *Nature Medicine*, 1–13-11–13 (2024).

62. Ikezogwo W*, et al.* Quilt-1m: One million image-text pairs for histopathology. *Advances in neural information processing systems* **36**, 37995-38017 (2023).

63. Zhang R, Weber C, Grossman R, Khan AA. Evaluating and interpreting caption prediction for histopathology images. In: *Proceedings of the 5th Machine Learning for Healthcare Conference* (eds Finale D-V*, et al.*). PMLR (2020).

64. Zhao Z, Kang H, Wang B, He C. Doclayout-yolo: Enhancing document layout analysis through diverse synthetic data and global-to-local adaptive perception. *arXiv preprint arXiv:241012628*, (2024).

65. Hu EJ*, et al.* Lora: Low-rank adaptation of large language models. *ICLR* **1**, 3-3 (2022).

66. Yu J, Wang Z, Vasudevan V, Yeung L, Seyedhosseini M, Wu Y. Coca: Contrastive captioners are image-text foundation models. *arXiv preprint arXiv:220501917*, (2022).

67. Lu MY, Williamson DFK, Chen TY, Chen RJ, Barbieri M, Mahmood F. Data-efficient and weakly supervised computational pathology on whole-slide images. *Nature biomedical engineering* **5**, 555–570-555–570 (2021).

68. Ilse M, Tomczak J, Welling M. Attention-based deep multiple instance learning. In: *International conference on machine learning*). PMLR (2018).

69. Cox DR. Regression models and life‐tables. *Journal of the royal statistical society: Series B (methodological)* **34**, 187-202 (1972).

## Acknowledgements

We extend special thanks to the teams at Ningbo Clinical Pathology Diagnosis Center, the First Affiliated Hospital of Soochow University, Affiliated Jinhua Hospital of Zhejiang University School of Medicine, Xiaogan Central Hospital of Wuhan University of Science and Technology, Hubei Maternal and Child Health Hospital, Guangdong Women and Children Hospital, and Hangzhou Adicon Clinical Laboratories for their invaluable support in data collection, annotation, and validation.

## Funding

This work was supported by the National Natural Science Foundation of China (grant no. 82430062 to T.G.), the Shenzhen Engineering Research Centre (grant no. XMHT20230115004 to Y.H.H.), the National Natural Science Foundation of China Youth Program (grant no. 62401329 to D.W.S.), the Tsinghua Shenzhen International Graduate School Cross-disciplinary Research and Innovation Fund Research Plan (grant no. JC2024002 to Y.H.H.), the Ningbo Top Medical and Health Research Program (grant no. 2023010211 to M.Z.), the Natural Science Foundation of Jiangsu Province (grant no. BK20241793 to L.C.G.), and the Suzhou Science and Technology Development Program (grant no. SKY2023009 to L.C.G.). Additional support was provided by Jilin FuyuanGuan Food Group Co., Ltd. through the joint research centre. C.H. acknowledges support from the University of Oxford and the Royal Society through a University Research Fellowship (URF/R1/241734).

## Author contributions:

Y.Z.W., L.C., T.G., Q.H., C.H., Y.H.H. and M.Z. conceptualized the study. Y.Z.W., L.C., T.G., Q.H., J.W.L., X.R.C., D.S.Y., Q.M.H., D.W.S. and Z.L. developed the methodology, while Y.Z.W., L.C., T.G., J.W.L., X.R.C., D.S.Y., X.T.L. and F.Y.Z. developed the software. Y.Z.W., L.C., T.G., Q.H., J.W.L., X.R.C., D.S.Y. and Z.L. curated the data, with Z.Y.S. contributing to data curation and scanner quality control; Y.Z.W., L.C., T.G., J.W.L., X.R.C. and Z.R.H. prepared the visualizations. X.D., B.H., H.Z.L., J.B.H., H.Y.G., Y.L., J.R.C., L.H.Z., Y.Q.L., Z.L. and F.Y.Z. performed validation, while Y.Z.W., L.C., T.G., Q.H., J.W.L., X.R.C., D.S.Y., Z.L. and L.C.G. performed formal analysis. Y.Z.W., L.C., Q.H., X.D., J.W.L., X.R.C., B.H., T.J.Z., Z.R.H., D.S.Y., Y.L., J.R.C., L.H.Z., Q.M.H., D.W.S., H.Z.L., J.B.H., H.Y.G. and Z.S. conducted the investigation; C.H., M.Z., Y.H.H., X.D., H.Z.L., J.B.H., H.Y.G., Y.L., X.L.Z. and B.H. provided resources. Y.H.H., C.H. and M.Z. acquired funding, and Y.H.H., C.H., M.Z. and L.C.G. administered and supervised the project. Y.Z.W., L.C., T.G., Q.H., J.W.L. and X.R.C. wrote the original draft; Y.H.H., C.H., M.Z., L.C. and T.G. reviewed and edited the manuscript. All authors reviewed and approved the final manuscript.

## Competing interests:

Authors declare that they have no competing interests.

Figures:

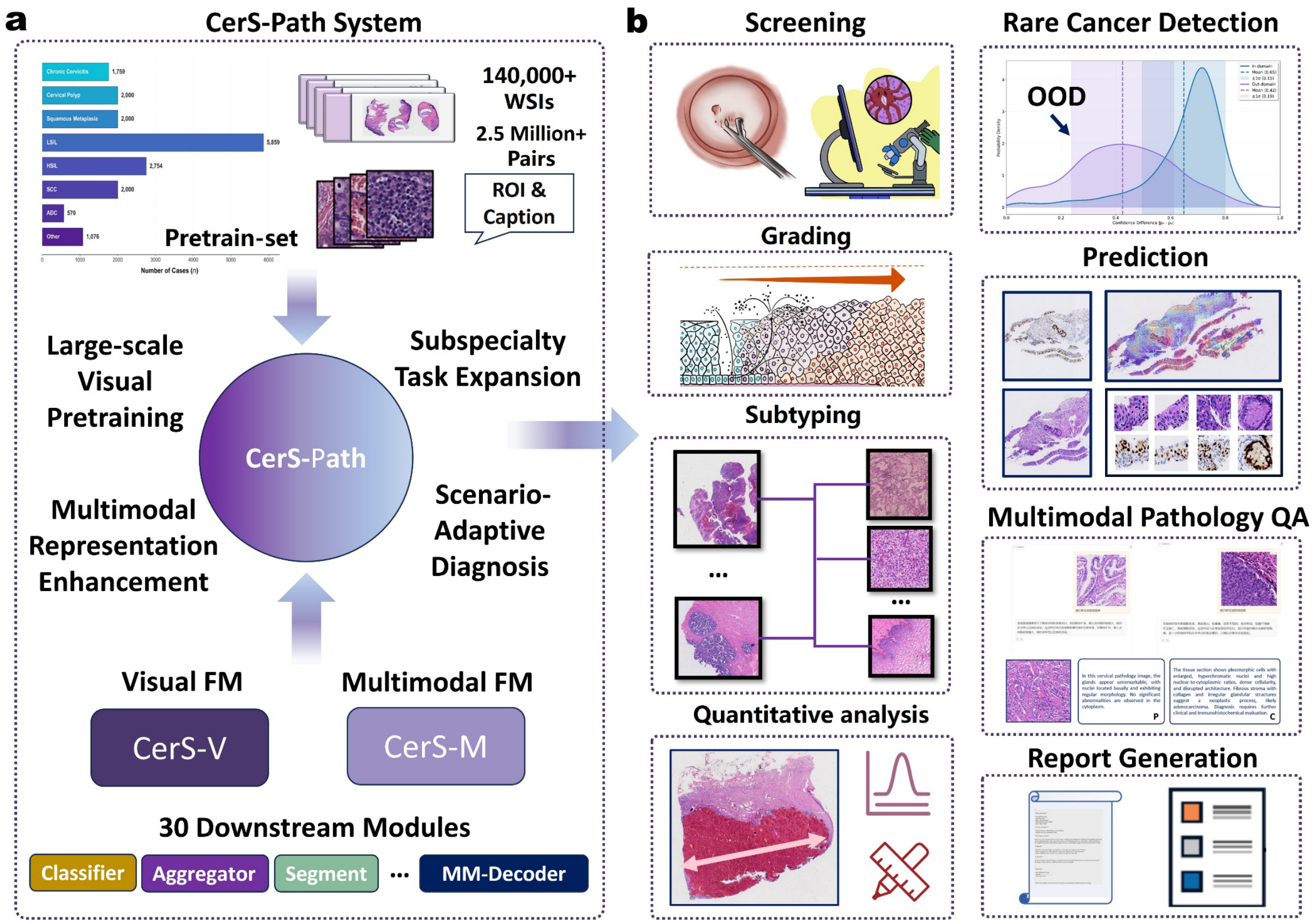

**Figure 1. Overview of the CerS-Path framework.** (**a**) CerS-Path architecture includes subspecialty pretraining data, subspecialty-specific foundation models, and modular diagnostic components. (**b**) Coverage of diagnostic functions supported by CerS-Path.

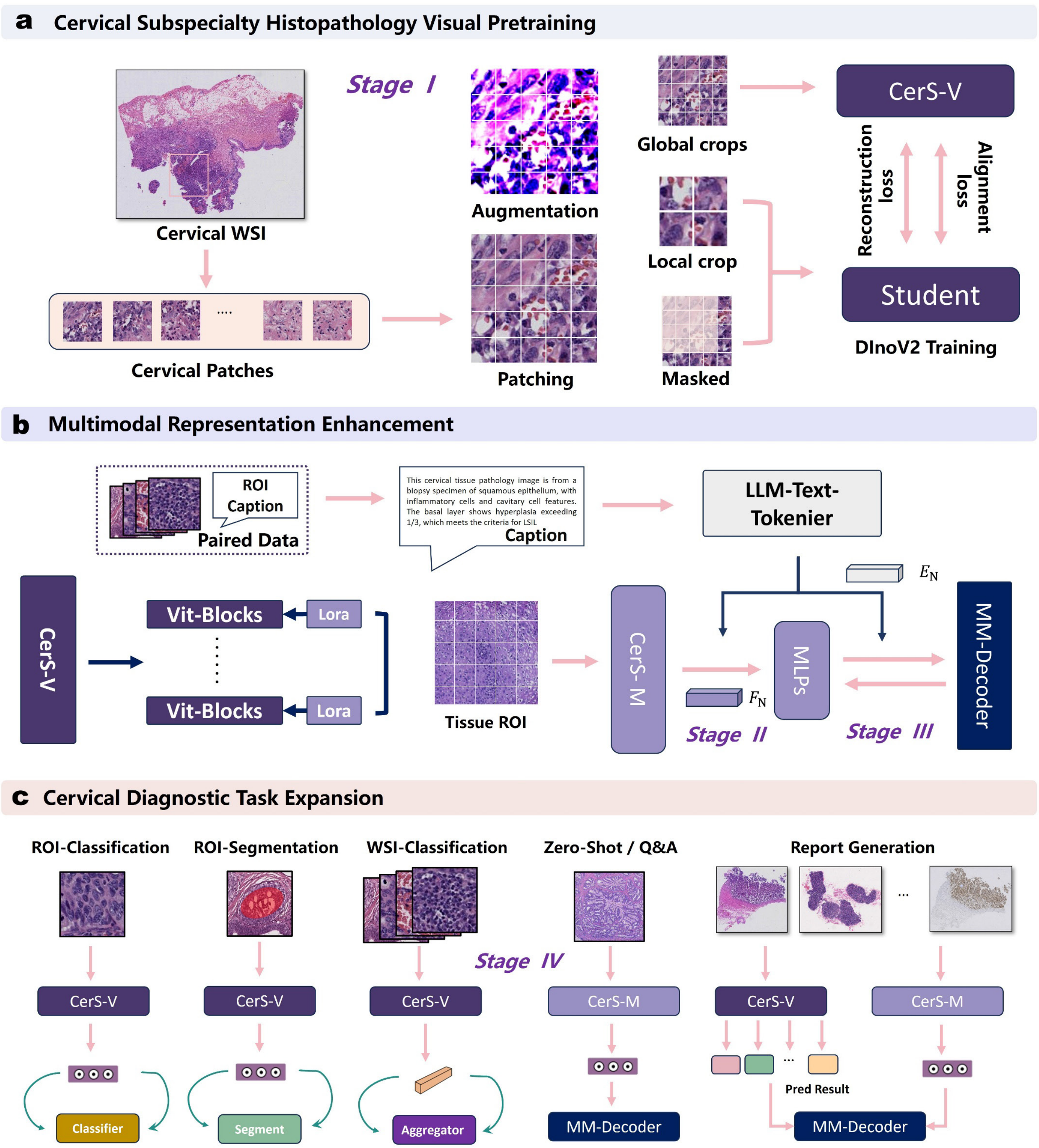

**Figure 2. Construction and functional overview of the CerS-Path system.** (**a**) Visual pretraining based on a curated subspecialty-level cervical pathology dataset to enhance visual representation capabilities. (**b**) Expansion of the subspecialty-specific multimodal representation space via multimodal mapping pretraining and instruction tuning. (**c**) Methods for extending downstream diagnostic functionalities.

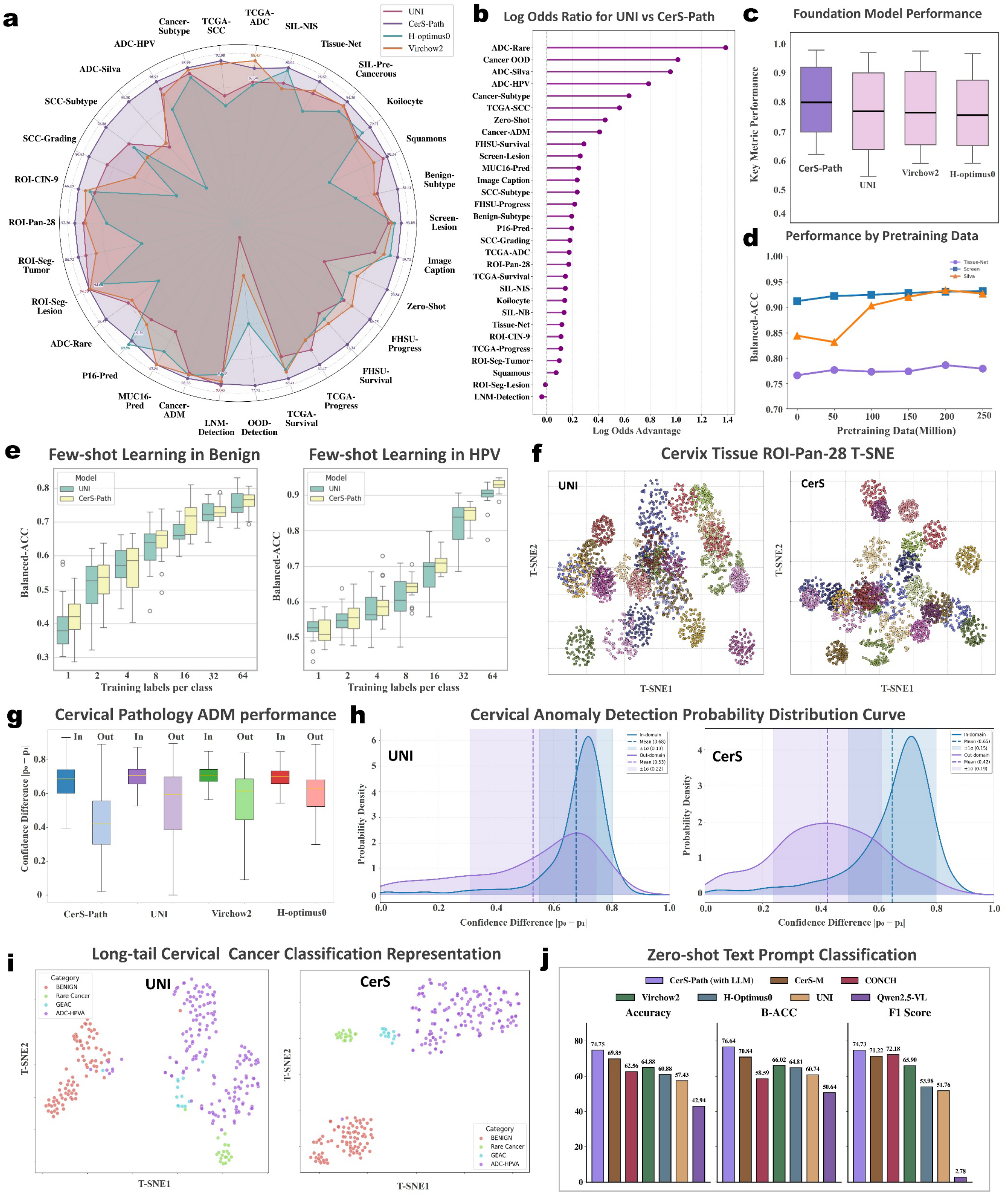

**Figure 3. Comprehensive evaluation of CerS-Path diagnostic performance and representation learning.** (**a**) Overview of the diagnostic performance of the CerS-Path system across all benchmark tasks; task abbreviations indicate diagnostic category and performance level. (**b**) Log odds ratios of CerS-Path over UNI across diagnostic tasks, indicating relative gains. (**c**) Foundational model performance comparison of CerS-Path versus UNI, Virchow2, and H-optimus0, where the boxes are statistics from 30 individual tasks. Detailed results for all 30 tasks

are provided in Supplementary Table 1. (**d**) Impact of pretraining sample size on foundational diagnostic performance for three representative tasks: Screen-Lesion, ADC-Silva, and Tissue-Net. (**e**) Comparison of few-shot learning results for CerS-Path. The x-axis shows the sample count per class, and 20 individual experiments are used to calculate the performance boxes. (**f**) T-SNE visualization of 28 cervical subspecialty tissue types. (**g**) Confidence distribution of in-domain and out-of-domain samples after training the open-set detection algorithm, with each box evaluated in 20 samples. (**h**) Out-of-domain sample detection via confidence scores. (**i**) T-SNE visualization of class distributions in long-tailed classification tasks. (**j**) Comparison of text-prompt zero-shot classification performance. For each box plot, the box extends from the 25th percentile (first quartile, Q1) to the 75th percentile (third quartile, Q3), with the central line indicating the median (50th percentile). Source data are provided as a Source Data file.

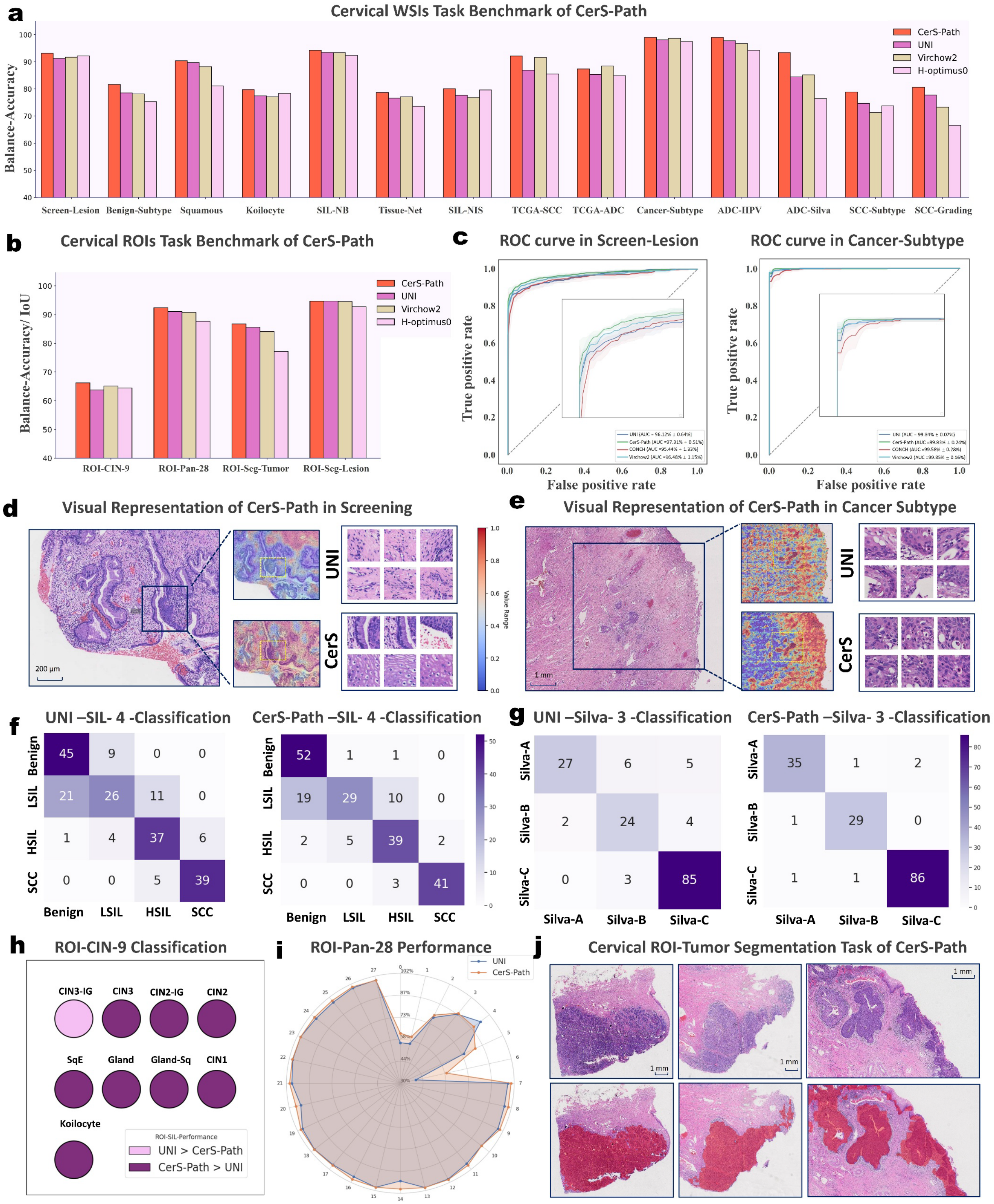

**Figure 4. CerS-Path performances on core visual diagnostic tasks.** (**a**) Benchmark for conventional visual tasks at the WSI level. (**b**) Benchmark for conventional visual tasks at the ROI level. (**c**) ROC curves for cervical screening tasks. Solid lines show mean ROC curves across five folds (*n*=5), with shaded bands indicating ±1 SD; AUC values are fold-wise means. (**d**) Visualization of micro-lesion detection in WSI screening. (**e**) Heatmaps and top-K patch visualization for difficult WSI subtyping cases. All heatmaps are representative of results observed

across 5-fold models from the validation cohort, all of which demonstrated similar attention patterns. (**f**) Confusion matrix of SIL grading. (**g**) Confusion matrix of ADC-Silva subtyping. (**h**) Per-class performance on ROI-CIN-9. (**i**) Classification comparison on ROI-Pan-28. (**j**) Segmentation results on ROI-Seg-Tumor. Source data are provided as a Source Data file.

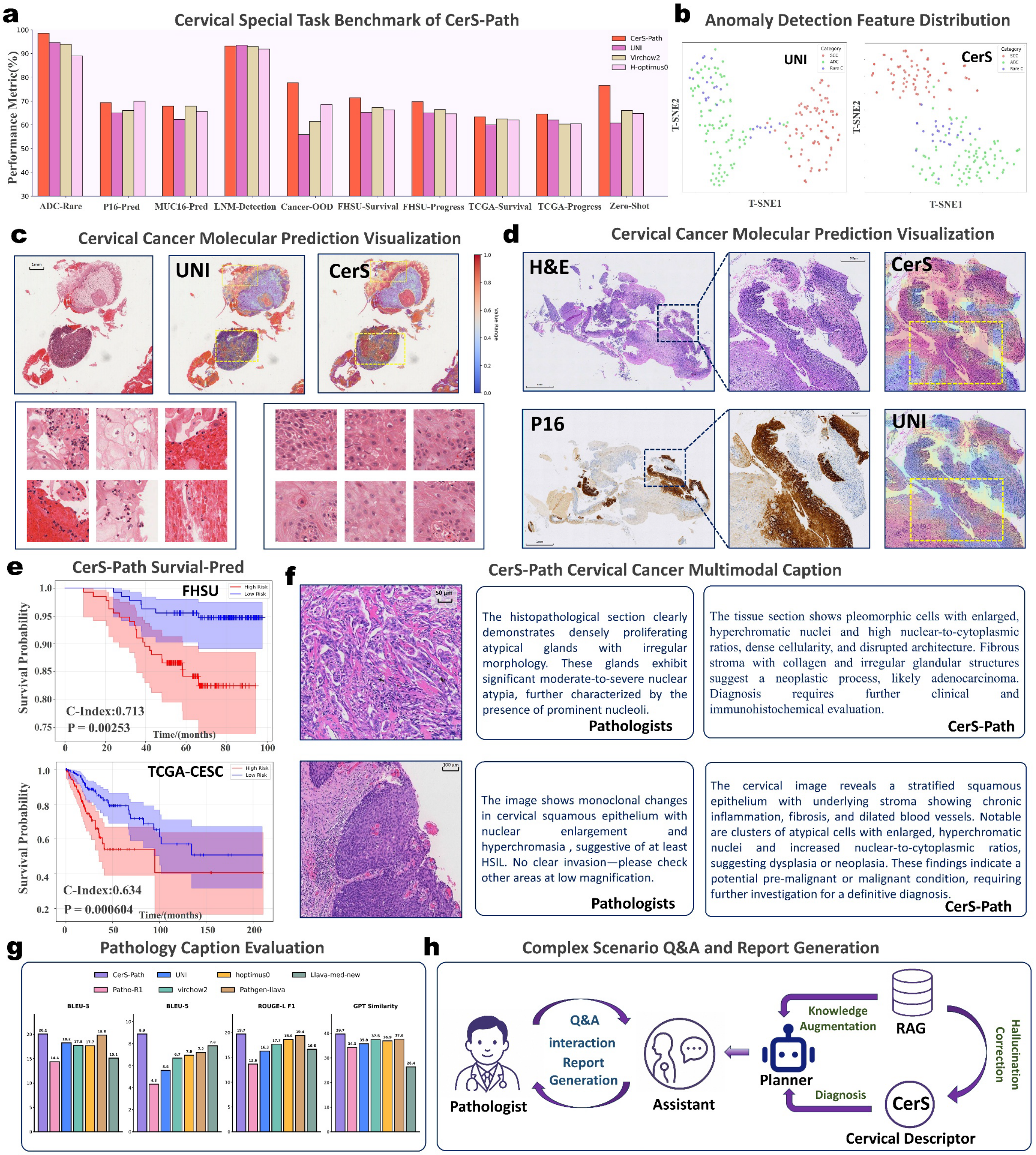

**Figure 5. Comprehensive evaluation of CerS-Path on specialized diagnostic tasks. (a)** Benchmark for subspecialty-specific diagnostic tasks. **(b)** Feature embedding in ADC-Rare, showing separation of GEAC and rare cancer classes under a long-tail distribution. **(c)** Visualization of model attention in MUC16 prediction on TCGA-CESC; differences highlighted between attention regions and Top-K patches. **(d)** Visualization of attention in P16 prediction task; CerS-Path attention maps compared with P16 immunostaining. **(e)** Kaplan-Meier curves of overall survival prediction based on CerS-Path risk stratification. Shaded bands indicate 95% confidence intervals estimated using the Greenwood formula. Differences between survival curves were compared using a two-sided log-rank test. No adjustments were made for multiple comparisons. The FHSU and TCGA-CESC cohorts included 268 and 259 patients, respectively, with high- and low-risk groups defined using the median predicted risk score. No adjustment was made for multiple comparisons. (**f**) Pathology image-based question-answering results from CerS-Path, compared with responses from pathologists. (**g**) Performance comparison of CerS-Path with other multimodal models on the Multimodal Pathology Captioning task. (**h**) Workflow of the diagnostic agent driven by a multimodal subspecialty foundation model. All heatmaps are representative of results observed across 5-fold models from the validation cohort, all of which demonstrated similar attention patterns. Source data are provided as a Source Data file.

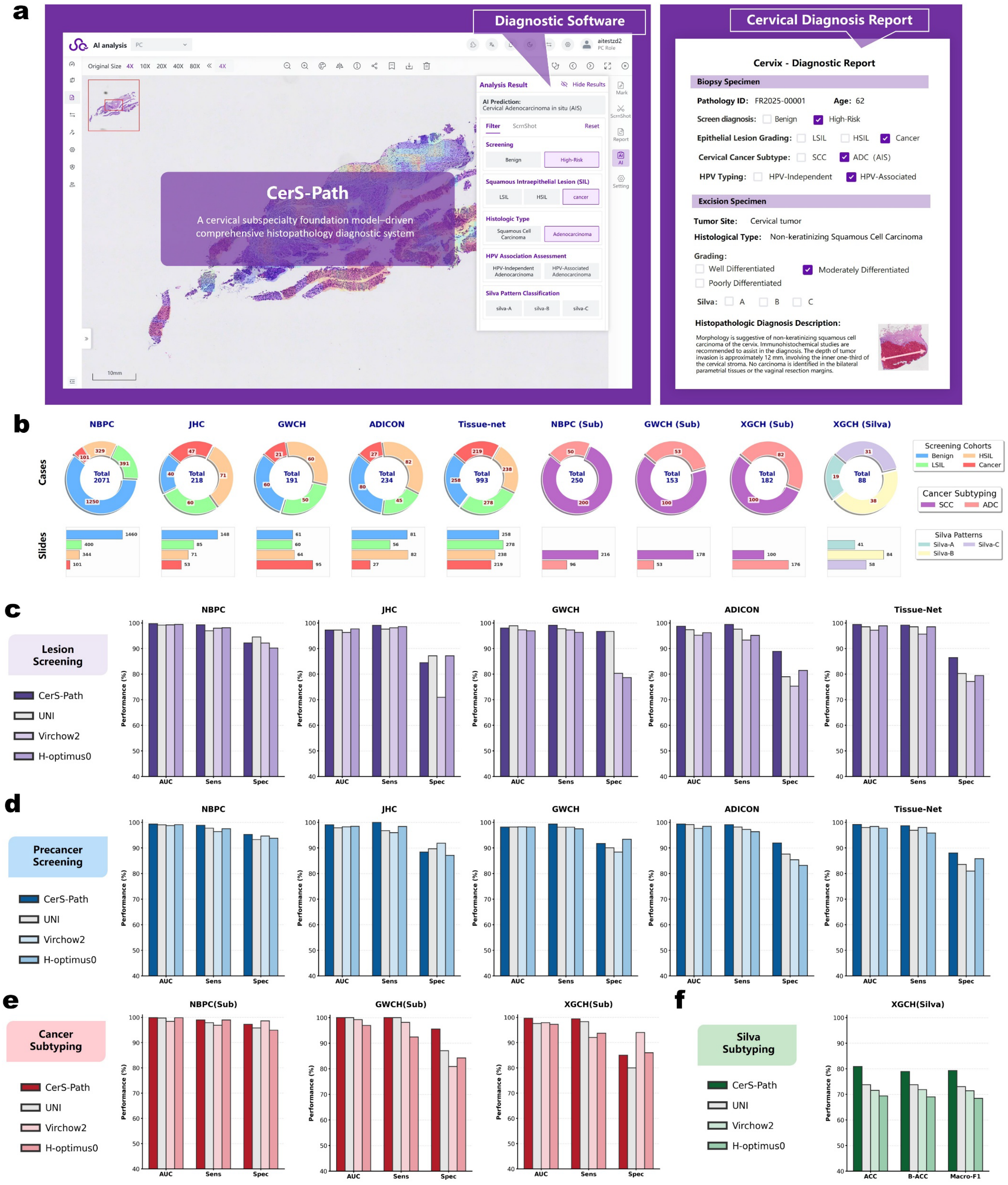

**Figure 6. Comprehensive prospective clinical evaluation of the CerS-Path system.** (**a**) Design of the CerS-Path diagnostic interface integrating core diagnostic functions. (**b**) Case distribution across participating clinical centers. **(c)** Performance of the lesion screening task in the multicenter validation cohort.**(d)** Performance of the precancer screening task in the multicenter validation cohort. **(e)** Performance of cancer subtyping in the multicenter validation cohort. (**f** ) Performance of Silva subtyping in the external validation cohort. Source data are provided as a Source Data file.